\documentclass[sigconf,nonacm]{acmart}

\AtBeginDocument{%
  \providecommand\BibTeX{{%
    \normalfont B\kern-0.5em{\scshape i\kern-0.25em b}\kern-0.8em\TeX}}}
\AtBeginDocument{%
  \providecommand\BibTeX{{%
    Bib\TeX}}}

\setcopyright{none}
\copyrightyear{}
\acmYear{}
\acmDOI{}
\acmISBN{}

\makeatletter
\def\authornotetext#1{
	\g@addto@macro\@authornotes{%
	\stepcounter{footnote}\footnotetext{#1}}%
}
\makeatother

\usepackage{multirow}
\usepackage{multicol}
\usepackage{booktabs}
\usepackage{subfigure}
\usepackage{subcaption}
\usepackage{hhline}
\usepackage{wasysym}
\usepackage{pifont}
\usepackage[ruled, vlined]{algorithm2e}
\SetKwInput{Require}{Require}
\SetKwInput{Ensure}{Ensure}
\usepackage{algpseudocode}
\usepackage{adjustbox}
\usepackage{graphicx}
\usepackage{balance}
\usepackage{caption}
\usepackage[most]{tcolorbox}
\usepackage{color, colortbl}
\usepackage{xcolor}
\usepackage{amsthm}
\usepackage{anyfontsize}
\usepackage{bbm}
\usepackage{enumitem}
\usepackage{xspace}
\tcbuselibrary{listings, breakable}

\newcommand{\model}{ScrapMem\xspace}
\renewcommand{\cite}{\citep}

\definecolor{RedOrange}{RGB}{255,69,0}

\definecolor{myup}{HTML}{A52A2A}

\definecolor{blue}{HTML}{C5D3E3}
\definecolor{rowcolor}{HTML}{D4FAFC}

\definecolor{PromptBlue}{RGB}{40, 60, 150}
\definecolor{CodeBg}{RGB}{248, 248, 248}
\newtcolorbox{promptbox}[1]{
    enhanced, colback=white, colframe=PromptBlue!80, arc=3mm, boxrule=1pt,
    left=4mm, right=4mm, top=5mm, bottom=3mm, fontupper=\small\rmfamily,
    attach boxed title to top left={xshift=5mm, yshift=-3mm, yshifttext=-1mm},
    boxed title style={colback=PromptBlue!10, colframe=PromptBlue, arc=2mm, boxrule=1pt},
    title={\color{black}\bfseries #1},
    before upper={\parskip=6pt}
}
\newtcolorbox{codeblock}{
    colback=CodeBg, colframe=gray!30, arc=1mm, boxrule=0.5pt,
    left=2mm, right=2mm, top=2mm, bottom=2mm,
    fontupper=\small\ttfamily, before upper={\parindent=0pt}
}
\renewcommand\footnotetextcopyrightpermission[1]{}
\acmConference[]{}{}{}

\begin{document}

\title{ScrapMem: A Bio-Inspired Framework for On-Device Personalized Agent Memory via Optical Forgetting}

\author{Jiale Chang}
\affiliation{%
  \department{College of Smart Agriculture}
  \institution{Nanjing Agricultural University}
  \city{Nanjing}
  \country{China}
}
\email{22423128@stu.njau.edu.cn}

\author{Yuxiang Ren}
\authornote{Corresponding author.}
\affiliation{
  \department{School of Intelligence Science and Technology}
  \institution{Nanjing University}
  \city{Suzhou}
  \country{China}
}
\email{renyuxiang@nju.edu.cn}

\renewcommand{\shortauthors}{Chang and Ren}
\begin{abstract}
Long-term personalized memory for LLM agents is challenging on resource-limited edge devices because multimodal histories grow continually and are costly to store. To address this challenge, we propose \model, a bio-inspired framework that consolidates each day's heterogeneous records into a unified ``Scrapbook Page.'' \model introduces \textit{Optical Forgetting}, an optical compression mechanism that progressively reduces the resolution and JPEG quality of older pages, lowering storage costs while keeping recent evidence at higher fidelity. To preserve semantic consistency and support structured retrieval, we construct an Episodic Memory Graph (EM-Graph) that organizes semantic nodes into causal-temporal EM-Paths. This design separates a stable retrieval structure from aging visual evidence, allowing relevant episodes to remain accessible as page fidelity decreases. Under the common ATM-Bench protocol, \model without forgetting establishes a new non-oracle state of the art with 51.0\% Joint@10 and reaches 70.3\% Recall@10. Optical Forgetting reduces the persistent memory footprint by up to 93.0\% relative to the raw corpus while retaining over 90\% of No-Forget performance. \model offers a storage-efficient and interpretable approach to long-term multimodal memory for on-device LLM agents. The code is available on \href{https://github.com/changjiale444/ScrapMem}{GitHub}.
\end{abstract}



\maketitle

\section{Introduction}
\begin{figure}[t]
  \centering
  \includegraphics[width=1.0\linewidth]{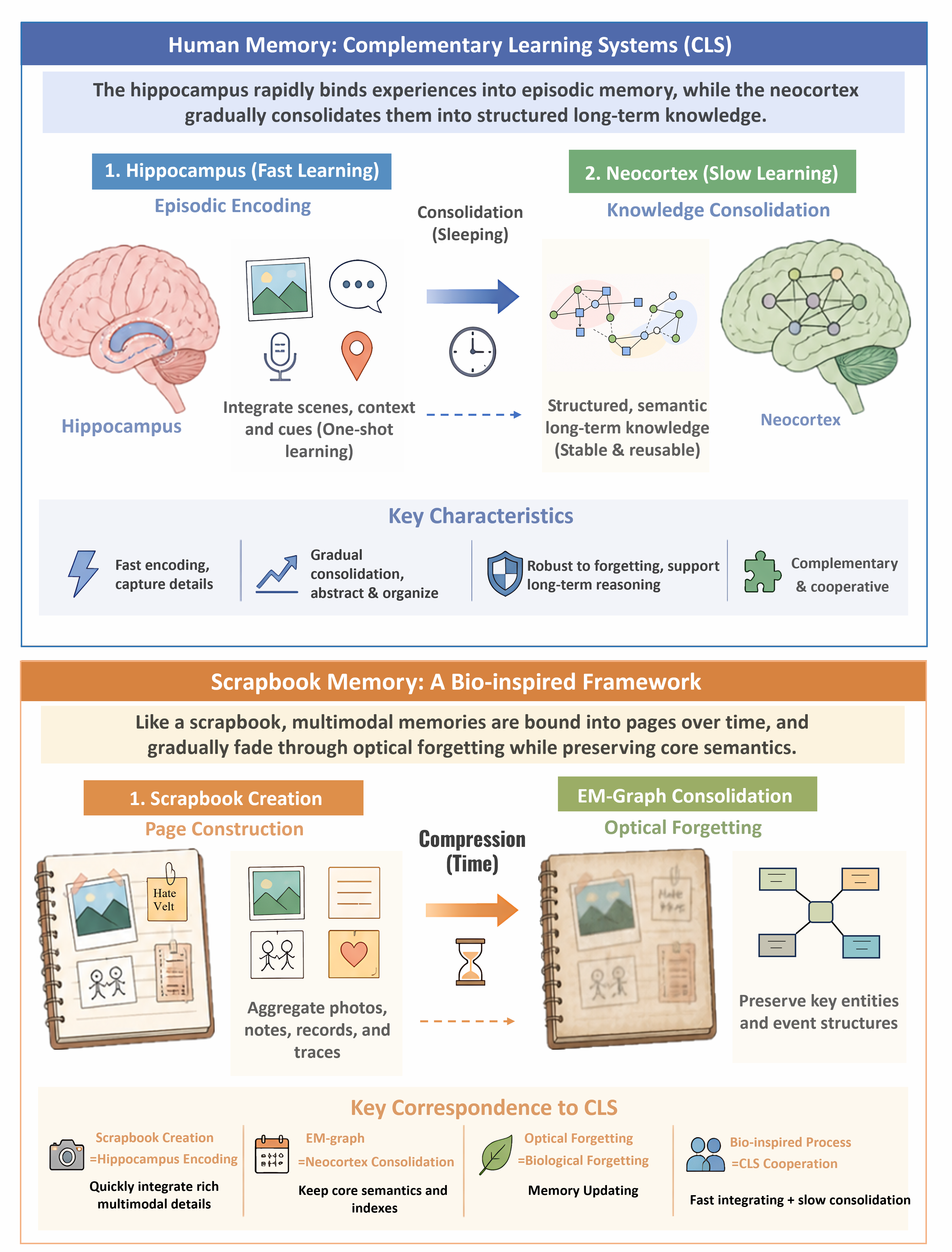}
  \Description{A conceptual comparison between Complementary Learning Systems and ScrapMem. Human episodic experiences are consolidated into long-term knowledge; ScrapMem consolidates daily multimodal records into visual pages, indexes them with an EM-Graph, and progressively compresses older pages.}
  \caption{Comparison between human memory (CLS theory) and Scrapbook Memory. Top: The hippocampus rapidly encodes multimodal episodic experiences, while the neocortex gradually consolidates them into stable long-term knowledge. Bottom: ScrapMem binds heterogeneous user data into Scrapbook Pages, indexes their event structure with the EM-Graph, and progressively compresses aging pages while retaining optically accessible evidence.}
  \label{fig:introduction}
\end{figure}

Personalized LLM agents must accumulate user-specific knowledge across interactions that may span months or years. Keeping this knowledge on edge devices is attractive for privacy and availability, but introduces strict storage and inference constraints~\cite{xu2024ondevice,yu2024cambricon,fu2024camphor}. Unlike document-centric knowledge bases, personal histories are inherently multimodal and multi-source: a single event may be recorded through an email, a photograph, a video clip, an application notification, and a later conversation~\cite{liu2025memverse,wu2024mobileagents,Meietal2026}. Answering a seemingly simple question can therefore require resolving personal references, locating evidence across sources, and relating events separated by long temporal intervals~\cite{maharana2024longterm,Meietal2026}. Directly replaying this growing history is constrained by model context windows and repeatedly incurs the cost of processing largely irrelevant records, making an external memory necessary~\cite{lewis2020rag,lee2024gist,lu2025multiturnrl}.

Existing agent memory systems primarily improve how textual memories are stored, updated, and retrieved. Some systems maintain flat memory banks and retrieve records by semantic similarity~\cite{borgeaud2022retro,zhong2024memorybank,chhikara2025mem0}; others construct linked notes or knowledge graphs to expose associations that are difficult to recover from isolated chunks~\cite{edge2024graphrag,gutierrez2024hipporag,xu2025amem}. These designs have substantially advanced long-term agent memory, but they do not directly resolve the representation problem in personalized multimodal settings. Cross-modal alignment remains difficult when evidence spans text, images, and time-stamped records~\cite{chen2025multimodalretrieval,liu2024retrievalreasoning,zhang2026trace}. Converting every record into text can discard visual evidence and layout, whereas retaining the original images and videos alongside textual indices causes the memory footprint to grow continually.

Recent advances in vision-language modeling make it possible to read dense visual layouts and combine textual and visual evidence within a shared inference process~\cite{bai2025qwen25vl,chen2024multimodalscaling}. Building on this capability, optical-context methods use images as dense carriers of textual history. DeepSeek-OCR maps long text contexts into two-dimensional visual representations~\cite{wei2025deepseekocr}; AgentOCR, MemOCR, and OCR-Memory further use rendered images to reduce the context or token cost of long-horizon agents~\cite{fengetal2026,shietal2026,lietal2026}. However, these methods primarily render textual trajectories for efficient model consumption. Personalized memory poses a broader representation problem: its inputs already contain native visual content, its evidence is distributed across sources and time, and its storage must remain manageable as the history ages. Together, these limitations leave a central gap: how can an on-device memory preserve native multimodal evidence, support structured temporal retrieval, and bound storage as a personal history grows?

As shown in Figure~\ref{fig:introduction}, we address this gap through complementary memory representations. Complementary Learning Systems (CLS) theory distinguishes rapid hippocampal encoding of episodic experience from gradual neocortical consolidation into structured knowledge~\cite{mcclelland1995cls,sun2023organizing}. CLS supplies the design abstraction: rapid event binding motivates daily visual pages, and structured consolidation motivates a persistent episodic index. Specifically, heterogeneous records occurring on the same day are consolidated into a \emph{Scrapbook Page}, which retains textual and visual evidence in a unified, temporally grounded unit. Across days, an Episodic Memory Graph (EM-Graph) connects extracted semantic nodes through event-centric EM-Paths, providing a structured index over the visual archive. The core design separates what is retained from how it is retrieved: visual pages preserve source evidence, while the EM-Graph maintains stable access paths across the growing history.

Long-term memory also requires a lifecycle policy. Fixed-capacity eviction creates a binary choice between retaining and deleting an old memory, while repeated textual summarization may alter the evidence on which later answers depend. We instead introduce \emph{Optical Forgetting}: as Scrapbook Pages age, their JPEG quality and spatial resolution are progressively reduced according to a temporal schedule. Recent pages retain fine-grained evidence, whereas older pages occupy less storage and preserve only the information that remains optically accessible. The EM-Graph is updated as pages degrade, retaining access to event structure while visually inaccessible nodes are pruned. Optical Forgetting therefore makes memory decay an explicit and controllable storage--fidelity trade-off within the same visual representation.

We instantiate these ideas in \model, a storage-efficient framework for on-device personalized agent memory. On ATM-Bench, which contains approximately four years of multimodal, multi-source personal history~\cite{Meietal2026}, \model without forgetting establishes a new non-oracle state of the art with 51.0\% Joint@10 and reaches 70.3\% Recall@10. With Optical Forgetting, the Timed-Gentle configuration reduces total storage from 4302.9 MiB of raw data to 299.5 MiB, a 93.0\% reduction, while retaining 46.3\% Joint@10. These results demonstrate that daily visual consolidation and structured episodic retrieval improve evidence-grounded personalized reasoning, while Optical Forgetting provides a controllable trade-off between persistent storage and evidence fidelity.

\textbf{Our contributions} are summarized as follows:
\begin{itemize}
    \item We propose \textit{ScrapMem}, which consolidates each day's heterogeneous personal records into a unified Scrapbook Page, preserving multimodal evidence in a temporally grounded visual memory unit.
    \item We introduce \textit{Optical Forgetting}, a deterministic progressive-degradation mechanism that exposes a controllable trade-off between the fidelity of aging memories and their storage footprint.
    \item We design an \textit{Episodic Memory Graph (EM-Graph)} that organizes semantic nodes into event-centric EM-Paths, providing a structured retrieval layer over the visually stored history.
    \item On ATM-Bench, \model (No-Forget) establishes a new non-oracle state of the art at 51.0\% Joint@10 and 70.3\% Recall@10, while Optical Forgetting provides a storage--quality trade-off with up to 93.0\% storage reduction. Complementary controlled experiments identify the resolution boundary and verify the respective effects of Optical Forgetting and the EM-Graph.
\end{itemize}

\section{Related Work}

\subsection{Long-Term Memory for Personalized Agents}

Long-term memory can be encoded parametrically or maintained in an external store~\cite{bai2025aimemory}. Parametric approaches internalize knowledge through continued pre-training, supervised fine-tuning, or parameter-efficient adaptation~\cite{brown2020gpt3,zheng2024llamafactory,hu2022lora}. They provide direct access through model weights, but repeated updates are costly and may cause catastrophic forgetting~\cite{luo2025forgetting}; hybrid approaches such as NextMem therefore explore compact latent factual memory~\cite{NextMem2026}. External memory preserves editable records outside the model and retrieves them on demand, making it better suited to continuously growing personal histories.

Early external-memory systems commonly store textual interaction records and retrieve them by recency, importance, or semantic relevance. Generative Agents combines these signals with reflection to support persistent simulated behavior~\cite{park2023generative}; MemoryBank introduces time- and importance-dependent updating for long-term conversational memory~\cite{zhong2024memorybank}; and MemGPT manages multiple memory tiers through explicit context-management operations~\cite{Packeretal2023}. More recent systems make memory organization itself adaptive. Mem0 extracts and updates compact user facts~\cite{chhikara2025mem0}, while A-MEM constructs interconnected notes whose attributes and links evolve as new memories arrive~\cite{xu2025amem}. These systems demonstrate that memory quality depends not only on retrieval, but also on how experiences are consolidated and maintained.

Graph-based memory further supports associative access beyond independent vector chunks. GraphRAG constructs entity-level communities for corpus-level summarization~\cite{edge2024graphrag}; HippoRAG and HippoRAG 2 use graph propagation to combine passage evidence with associative memory~\cite{gutierrez2024hipporag,gutierrez2025hipporag2}; and human-inspired episodic memory organizes long sequences into retrievable events with temporal contiguity~\cite{Fountas2025}. ScrapMem shares the motivation of retaining relational and episodic structure, but addresses a different representation setting. Its memory items originate from multimodal personal records, and its EM-Graph indexes temporally grounded visual pages while preserving the underlying evidence alongside textual summaries.

\subsection{Multimodal Personalized Memory}

Personalized memory differs from conventional document retrieval because references are user-dependent and evidence may be distributed across modalities and sources. ATM-Bench formalizes this setting with approximately four years of emails, images, and videos, including questions that require personal-reference resolution and multi-evidence reasoning~\cite{Meietal2026}. MemVerse similarly argues for lifelong multimodal memory and combines hierarchical retrieval with structured long-term representations~\cite{liu2025memverse}. Meanwhile, edge-oriented multimodal systems expose the practical cost of retaining and repeatedly processing visual histories~\cite{cai2024recall,li2025venus}, while trusted-execution frameworks such as AgenTEE highlight the privacy motivation for keeping agent state local~\cite{AgenTEE2026}. Together, these studies motivate a memory design that preserves native multimodal evidence while controlling long-term storage growth.

ScrapMem treats the day as the basic unit of consolidation. Records from the same day are placed on a shared Scrapbook Page, preserving temporal co-occurrence and cross-modal context before semantic extraction. The page remains available as source evidence, while its extracted nodes and EM-Paths provide a lightweight retrieval structure. This evidence--index separation distinguishes ScrapMem from approaches that retain only textual summaries or embedding-level representations of the original experience.

\subsection{Optical Representations for Agent Memory}

Optical context compression maps one-dimensional text into a two-dimensional visual carrier, allowing a VLM to process more information within a limited visual-token budget. DeepSeek-OCR demonstrates the feasibility of this paradigm for long contexts~\cite{wei2025deepseekocr}. AgentOCR renders observation--action trajectories and introduces segment-level optical caching and learned self-compression~\cite{fengetal2026}. MemOCR uses structured rich-text layouts and budget-aware optimization to allocate visual space to important content~\cite{shietal2026}. OCR-Memory stores rendered trajectories with visual identifiers and retrieves evidence through a locate-and-transcribe procedure~\cite{lietal2026}. These methods establish optical memory as an alternative to repeatedly serializing long textual histories.

ScrapMem extends the optical-memory perspective in two directions. First, the visual page serves as a native container for heterogeneous personal records, including images, visual evidence extracted from videos, and textual content. Second, Optical Forgetting assigns age-dependent resolution and JPEG quality to the stored pages, producing recency-graded fidelity without training an additional compression policy. The EM-Graph complements this physical memory layer by retaining event-centric access paths as page fidelity decreases.

\subsection{Forgetting and Memory Consolidation}

Forgetting has been implemented at several levels in existing memory systems. MemoryBank decays the retrievability of textual memories over time and reinforces memories when they are recalled~\cite{zhong2024memorybank}. Parametric approaches address forgetting during continual model updates~\cite{luo2025forgetting}, whereas structured memories consolidate or revise stored representations as new information arrives~\cite{xu2025amem,liu2025memverse}. These mechanisms primarily alter retrieval scores, learned parameters, or semantic records. Optical Forgetting applies deterministic lossy compression directly to the stored visual evidence, translating recency-dependent memory fidelity into an interpretable storage policy. The biological connection serves as a design abstraction for combining daily consolidation, structured retrieval, and long-term storage management.

\section{Preliminaries}

In this section, we formalize long-term personalized referential memory QA as the task for on-device multimodal agent memory.

We consider long-term personalized memory to include a user’s long-term, multimodal, and multi-source personal memories. We denote the multimodal personal memory corpus as \(\mathcal{D}\), which consists of three types of raw user data:

\begin{equation}
\mathcal{D}=\left\{I_{i}\right\}_{i=1}^{N_{I}} \cup\left\{V_{i}\right\}_{i=1}^{N_{V}} \cup\left\{T_{i}\right\}_{i=1}^{N_{T}},
\label{eq:memory_corpus}
\end{equation}

where \(\{I_{i}\}\) denotes image memories, \(\{V_{i}\}\) denotes video memories, and \(\{T_{i}\}\) denotes textual memories.

To obtain a unified representation from the raw personal memory data \(\mathcal{D}\), we define a memory construction function \(\mathcal{U}\) that transforms the heterogeneous corpus into a standardized memory store \(\mathcal{M}\):
\begin{equation}
\mathcal{M} = \mathcal{U}(\mathcal{D}),
\label{eq:memory_store}
\end{equation}
where \(\mathcal{M}\) serves as a generic and unified memory representation for all downstream retrieval and reasoning procedures.

Given a query \(q\), the agent retrieves a set of relevant memory items 
\(\mathcal{E} \subseteq \mathcal{M}\) using a retriever \(\mathcal{R}\), 
and generates an answer \(\hat{a}\) based on the retrieved evidence:

\begin{equation}
\mathcal{E} = \mathcal{R}(q, \mathcal{M},k),
\label{eq:retrieval}
\end{equation}
where \(\mathcal{E} \subseteq \mathcal{M}\) denotes the retrieved evidence set.

An answerer \(\mathcal{A}\) then generates an answer conditioned on the query and retrieved evidence:
\begin{equation}
\hat{a} = \mathcal{A}(q, \mathcal{E}).
\label{eq:answer_gen}
\end{equation}

To model realistic long-term memory dynamics, we further define two cognitive operations in a generic form:

\textbf{1. Memory Linking.}
We define a memory linking function that establishes relational connections between memory items:
\begin{equation}
\mathcal{L}: \mathcal{M} \to \mathbb{R}^{N \times N},
\label{eq:memory_link}
\end{equation}
where \(N\) is the number of memory items in \(\mathcal{M}\), and \(\mathbb{R}^{N \times N}\) denotes the adjacency matrix of relational connections.

\textbf{2. Memory Forgetting.}
We define a memory forgetting function that updates memory states over time:
\begin{equation}
\mathcal{F}: \mathcal{M}_t \to \mathcal{M}_{t+\tau},
\label{eq:memory_forget}
\end{equation}
which models changes in memory accessibility over time.

\section{Method: \model}

\begin{figure*}[t]
  \centering
  \includegraphics[width=1.0\linewidth]{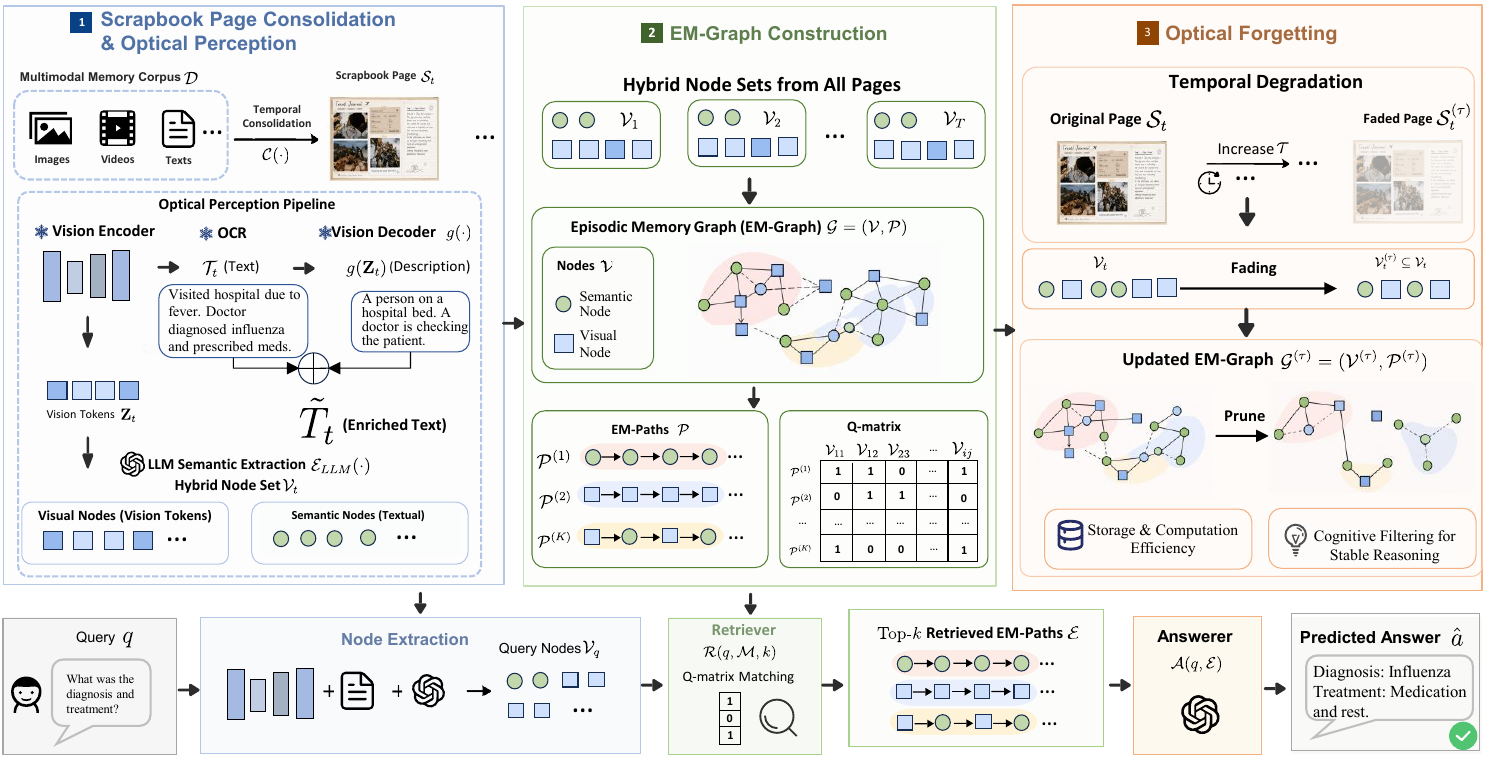}
  \Description{The ScrapMem pipeline has three stages: daily multimodal records are consolidated and perceived as Scrapbook Pages; semantic and visual nodes are organized into EM-Paths and an EM-Graph; and aging pages are compressed through Optical Forgetting while the graph is updated. Queries retrieve relevant paths and source pages for answer generation.}
    \caption{
Overview of ScrapMem. 
(1) \textbf{Consolidation and Perception}: Consolidates heterogeneous records (images, videos, and text) into daily Scrapbook Pages and extracts hybrid representations through OCR and visual-semantic perception. 
(2) \textbf{EM-Graph Construction}: Organizes nodes into an Episodic Memory Graph with event-centric paths (EM-Paths) for structured retrieval and multi-hop reasoning. 
(3) \textbf{Optical Forgetting}: Progressively compresses aging pages to reduce storage while retaining optically accessible evidence. 
The system retrieves relevant EM-Paths to generate personalized, evidence-grounded responses.
}
  \label{fig:ScrapMem1}
\end{figure*}

Figure~\ref{fig:ScrapMem1} presents the three stages of ScrapMem: Scrapbook Page Consolidation and Optical Perception, EM-Graph Construction, and Optical Forgetting. Together, these stages preserve multimodal evidence, construct a structured retrieval index, and manage the storage cost of aging memories.
\subsection{Scrapbook Page Consolidation and Optical Perception}

Given the multimodal memory corpus \(\mathcal{D}\) defined in Eq.~(\ref{eq:memory_corpus}), we first organize heterogeneous user data into temporally grounded \emph{Scrapbook Pages}, which serve as the basic units of episodic memory. We use one calendar day as the consolidation interval and aggregate all multimodal records from that day into a unified page representation:
\begin{equation}
\mathcal{S}_t = \mathcal{C}(\{I_i, V_j, T_k\}_{t}),
\end{equation}
where \(\mathcal{C}(\cdot)\) is the temporal consolidation function and $\mathcal{S}_t$ is the Scrapbook Page for day $t$.

Each Scrapbook Page preserves visual content and textual information in a common layout. An \emph{optical perception} pipeline then converts the page into representations suitable for semantic indexing. The vision encoder processes the rendered page to produce a sequence of visual tokens \(\mathbf{Z}_t\), while an OCR module produces the recognized text \(T_t\).

We further map the visual tokens to textual descriptions through a vision-to-text transformation:
\begin{equation}
\tilde{T}_t = T_t \oplus g(\mathbf{Z}_t),
\end{equation}
where \(g(\cdot)\) denotes a semantic description function, such as image captioning, and \(\oplus\) denotes concatenation.

An LLM extracts nodes from \(\tilde{T}_t\) and from the original text records \(\mathcal{T}_t=\{T_k\}_t\):
\begin{equation}
\mathcal{V}_t = \mathcal{X}_{LLM}(\tilde{T}_t) \cup \mathcal{X}_{LLM}(\mathcal{T}_t),
\end{equation}
where \(\mathcal{X}_{LLM}(\cdot)\) extracts entities, events, and relations. Nodes from optical perception and original text are linked into EM-Paths, while Scrapbook Pages retain the multimodal evidence (see Appendix~\ref{sec:appendix_prompts} for prompts).

\subsection{EM-Graph Construction}

To enable structured reasoning over long-term memories, we organize the extracted node sets into an Episodic Memory Graph (EM-Graph). The EM-Graph is defined as an event-structured graph \(\mathcal{G} = (\mathcal{V}, \mathcal{P})\), where \(\mathcal{V} = \bigcup_t \mathcal{V}_t\) denotes the global node set and \(\mathcal{P}\) denotes the set of all episodic memory paths (see Algorithm~\ref{alg:node_merge} in Appendix~\ref{sec:appendix_algorithms}).

Within each Scrapbook Page, related nodes are arranged into \emph{Episodic Memory Paths (EM-Paths)}, which capture local causal-temporal event chains. Each EM-Path is an ordered subset of nodes originating from the same page:
\begin{equation}
P_t^{(r)} = \big\{v_{t,1}, v_{t,2}, \dots, v_{t,\ell_r}\big\},\quad v_{t,i} \in \mathcal{V}_t,
\end{equation}
where \(r\) indexes distinct semantic paths within page \(t\). EM-Paths encode coherent event-centric sequences (e.g., \emph{diagnosis} \(\rightarrow\) \emph{treatment} \(\rightarrow\) \emph{recovery}) and serve as the retrieval units for multi-hop reasoning.

To facilitate efficient matching during retrieval, we explicitly encode node--path relationships via a binary incidence matrix (Q-matrix):
\begin{equation}
Q \in \{0,1\}^{|\mathcal{P}| \times |\mathcal{V}|}, \quad
Q_{ij} =
\begin{cases}
1, & v_j \in P_i, \\
0, & \text{otherwise},
\end{cases}
\end{equation}
where \(\mathcal{P} = \{P_i\}\) denotes the set of all EM-Paths. This structure enables efficient matching between query nodes and candidate paths by measuring node overlap.

For retrieval, we apply query-side semantic extraction to an input query \(q\) and obtain a set of query nodes. Following the problem formulation in Preliminaries, the retriever \(\mathcal{R}\) uses the node--path incidence matrix \(Q\) to measure overlap between matched graph nodes and candidate paths, then returns the top-\(k\) EM-Paths as memory evidence:
\begin{equation*}
\mathcal{E} = \mathcal{R}(q, \mathcal{M}, k),
\end{equation*}
where \(\mathcal{E} \subseteq \mathcal{M}\) denotes the retrieved evidence set consisting of relevant episodic memory paths. The retrieved evidence \(\mathcal{E}\) is then fed into the answerer \(\mathcal{A}\) for downstream reasoning and response generation, producing the final predicted answer \(\hat{a} = \mathcal{A}(q, \mathcal{E})\). On ATM-Bench, we rerank the fine-grained source records within the pages linked to the retrieved EM-Paths. The top ten records serve as input to the VLM and are used to compute Recall@10.

\subsection{Optical Forgetting}

To manage long-term memory under a finite storage budget, \emph{Optical Forgetting} progressively compresses aging Scrapbook Pages.

For a Scrapbook Page created at time \(t\), let \(\tau\) denote its age at the current update. We formulate forgetting as a time-dependent lossy transformation:
\begin{equation}
\mathcal{S}_t^{(\tau)} = \mathcal{F}_{\tau}(\mathcal{S}_t),
\end{equation}
where \(\mathcal{F}_{\tau}(\cdot)\) applies the resolution scale and JPEG quality associated with age \(\tau\). The transformation reduces storage while progressively lowering the legibility of fine-grained visual and textual details.

As compression intensifies, fine-grained clues become less accessible to optical perception and semantic-node extraction. Re-perception of the degraded page therefore yields a subset of the original nodes:
\begin{equation}
\mathcal{V}_t^{(\tau)} \subseteq \mathcal{V}_t,
\end{equation}
where \(\mathcal{V}_t^{(\tau)}\) contains the nodes that remain reliably perceptible after compression.

The EM-Graph is updated accordingly. Nodes that can no longer be reliably perceived are removed, and EM-Paths that lose semantic coherence are pruned. The updated graph is
\begin{equation}
\mathcal{G}^{(\tau)} = \big(\mathcal{V}^{(\tau)}, \mathcal{P}^{(\tau)}\big),
\end{equation}
where \(\mathcal{V}^{(\tau)}\) is the pruned global node set and \(\mathcal{P}^{(\tau)}\) is the set of retained EM-Paths. The updated graph contains only the structure supported by the remaining page evidence and cannot reconstruct details removed by compression.

Optical Forgetting therefore serves two roles. It bounds persistent storage by compressing aging pages, and it keeps the EM-Graph consistent with evidence that remains optically accessible (see Algorithm~\ref{alg:pruning} in Appendix~\ref{sec:appendix_algorithms}). The mechanism does not reconstruct information removed by compression; instead, it exposes a controllable trade-off between page fidelity and storage footprint.

\section{Experiments and Results}

\begin{table*}[t]
\centering
\small
\definecolor{ReferenceGreen}{RGB}{234,246,236}
\definecolor{OursPurple}{RGB}{242,238,252}
\begin{tabular}{cl c cccccc}
\toprule
\multirow{2}{*}{\textbf{\#}} & \multirow{2}{*}{\textbf{System}} & \textbf{Memory} & \multicolumn{6}{c}{\textbf{ATM-Bench}} \\
\cmidrule(lr){3-3} \cmidrule(lr){4-9} 
& & \textbf{Rep.} & \textbf{QS} & \textbf{R@10} & \textbf{Joint@10} & \textbf{N} & \textbf{R} & \textbf{O} \\
\midrule

\multicolumn{9}{l}{\textit{Upper / Lower Bounds}} \\
\midrule
\rowcolor{ReferenceGreen}
1 & No-Evidence & -- & 0.2 & -- & -- & 0.0 & 0.0 & 0.6 \\
\rowcolor{ReferenceGreen}
2 & Oracle & DM & 70.0 & -- & -- & 81.8 & 69.3 & 61.9 \\
\rowcolor{ReferenceGreen}
3 & Oracle & SGM & 77.8 & -- & -- & 85.0 & 90.4 & 69.5 \\
\midrule

\multicolumn{9}{l}{\textit{Memory Agents}} \\
\midrule
4 & A-Mem (Piled) & DM & 46.1 & 66.6 & 44.0 & 55.0 & 27.0 & 44.9 \\
5 & A-Mem (Linked) & DM & 44.8 & 66.4 & 42.8 & 57.2 & 16.9 & 43.6 \\
6 & Mem0$_{\textit{Agentic}}$ & DM & 43.5 & 61.9 & 41.8 & 57.2 & 25.9 & 48.4 \\
7 & Mem0$_{\textit{Plain}}$ & DM & 41.5 & 61.9 & 38.4 & 58.6 & 25.9 & 33.7 \\
\midrule

\multicolumn{9}{l}{\textit{RAG Systems}} \\
\midrule
8 & HippoRAG2 & DM & 42.9 & 66.4 & 41.5 & 58.6 & 34.9 & 34.1 \\
9 & HippoRAG2 & SGM & 47.7 & 69.6 & 46.9 & 59.4 & 39.3 & 41.8 \\
10 & Self-RAG & DM & 42.1 & 61.8 & 42.8 & 46.4 & 33.1 & 30.9 \\
11 & Self-RAG & SGM & 50.3 & 68.7 & 48.2 & 59.7 & 35.3 & 48.4 \\
12 & ATM-RAG & DM & 42.0 & 61.8 & 41.3 & 50.6 & 31.0 & 38.9 \\
13 & ATM-RAG & SGM & 51.0 & 68.7 & 48.6 & \textbf{60.3} & 32.4 & 48.2 \\
\midrule

\multicolumn{9}{l}{\textit{Ours}} \\
\midrule
\rowcolor{OursPurple}
14 & \textbf{\model (No-Forget)} & Graph & \textbf{52.5} & \textbf{70.3} & \textbf{51.0} & 57.2 & \textbf{50.4} & \textbf{48.8} \\
\rowcolor{OursPurple}
15 & \textbf{\model (Opt.-Forget)} & Graph & 48.4 & 66.1 & 46.9 & 55.3 & 45.3 & 42.8 \\
\bottomrule
\end{tabular}
\caption{ATM-Bench performance with top-10 evidence. Light-green rows are bounds (No-Evidence lower; Oracle DM/SGM upper), whereas light-purple rows are \model. QS denotes Question Type Score; N/R/O denote Number/List Recall/Open-ended. \model (No-Forget) leads all non-oracle systems in Joint@10 and R@10.}
\label{tab:main_results}
\end{table*}

\subsection{Experimental Setup}

\noindent\textbf{Dataset.}
We evaluate \model on ATM-Bench~\cite{Meietal2026}, which contains approximately four years of privacy-preserving records: 6,741 emails, 3,759 images, and 533 videos. Its 1,038 question--answer--evidence instances are human annotated and independently validated for evidence grounding, completeness, and temporal consistency. The benchmark therefore tests both answering and evidence retrieval over a long, multi-source history.

\noindent\textbf{Metrics.}
Following ATM-Bench~\cite{Meietal2026}, we evaluate both answer correctness and retrieval quality. ATM-Bench defines three answer categories: \texttt{Number}, \texttt{List Recall}, and \texttt{Open-ended}. Given a question $q$, ground-truth answer $a$, predicted answer $\hat{a}$, and question type $t$, the Question Type Score is
{\small
\begin{equation}
\mathrm{QS}(q,a,\hat{a}) =
\begin{cases}
\mathrm{EM}(a,\hat{a}), & t = \texttt{Number},\\[1pt]
J(a,\hat{a}), & t = \texttt{List Recall},\\[1pt]
\mathrm{LLM\text{-}Judge}(q,a,\hat{a}), & t = \texttt{Open-ended}.
\end{cases}
\label{eq:qs}
\end{equation}
}
The \textbf{N}, \textbf{R}, and \textbf{O} columns in Table~\ref{tab:main_results} report these three category-specific scores, respectively. In particular, the standalone \textbf{R} denotes the List Recall answer score, not retrieval recall.

Number questions use Exact Match, whereas List Recall questions use Jaccard overlap:
\begin{gather}
\mathrm{EM}(a,\hat{a}) =
\begin{cases}
1, & a=\hat{a},\\
0, & \text{otherwise},
\end{cases}
\label{eq:em}\\[2pt]
J(a,\hat{a}) =
\frac{
| \texttt{toList}(a)\cap \texttt{toList}(\hat{a}) |
}{
| \texttt{toList}(a)\cup \texttt{toList}(\hat{a}) |
}.
\label{eq:jaccard}
\end{gather}
Open-ended answers are evaluated by the benchmark's binary LLM judge.

Given the gold evidence set $E$ and top-$k$ retrieved evidence $\hat{E}_k$, retrieval quality and the joint score are
\begin{gather}
\mathrm{Recall@}k =
\frac{|E \cap \hat{E}_k|}{|E|},
\label{eq:recall}\\[2pt]
\mathrm{Joint@}k =
\mathrm{QS}(q,a,\hat{a}) \times \mathrm{Recall@}k.
\label{eq:joint}
\end{gather}
We report $k=10$ throughout. Thus, a high Joint@10 requires both a correct answer and retrieval of its supporting evidence.

\noindent\textbf{Baselines.}
We follow the official configurations for memory agents A-Mem~\cite{xu2025amem} and Mem0~\cite{chhikara2025mem0}, and retrieval systems HippoRAG2~\cite{gutierrez2025hipporag2}, Self-RAG~\cite{asai2024selfrag}, and ATM-RAG~\cite{Meietal2026}. Descriptive Memory (DM) stores records as text, whereas Schema-Guided Memory (SGM) uses fields such as time, location, entities, OCR, and tags. Piled variants retrieve independent records; Linked variants additionally organize relations among them. Table~\ref{tab:main_results} includes the official organization and answering variants, with No-Evidence and Oracle as lower and upper references.

\noindent\textbf{Implementation Details.}
Memory processing uses Qwen3-VL-2B-Instruct~\cite{bai2025qwen3vl}; retrieval uses all-MiniLM-L6-v2~\cite{reimers2019sentencebert}; and answering uses Qwen3-VL-8B-Instruct. All systems share the top-$k{=}10$ evidence budget, answer prompt, and decoding configuration; videos contribute at most eight frames. GPT-5-mini judges only Open-ended answers. Appendix~\ref{sec:appendix_setup} gives the remaining settings and threshold sweep.

\noindent\textbf{Evaluation Scope.}
The ATM-Bench evaluation first compares end-to-end evidence-grounded answering, then examines the forgetting schedule, canonical threshold, and storage--performance trade-off. Section~\ref{sec:controlled_main} complements this public-benchmark evaluation with controlled experiments on synthetic personal histories under a recency-aware filtering objective; Appendix~\ref{sec:controlled_synthetic} describes their construction and protocol.

\subsection{Main Results}
Table~\ref{tab:main_results} is our main experiment: an end-to-end comparison with memory-agent and RAG baselines on public ATM-Bench under a shared top-10 protocol. No-Forget evaluates the uncompressed system; Opt.-Forget evaluates the complete storage-aware pipeline. We then analyze overall performance, answer types, retrieval, and the fidelity cost of Optical Forgetting.

\noindent\textbf{Overall Performance.}
Among non-oracle systems, \model (No-Forget) achieves the highest QS (52.5\%), R@10 (70.3\%), and Joint@10 (51.0\%). It exceeds the strongest RAG baseline, ATM-RAG-SGM (48.6\%), by 2.4 Joint@10 points under the same top-10 budget. Improving both QS and R@10 matters because Joint@10 penalizes either a retrieval or answering failure. The result is consistent with the framework's division of labor: daily pages retain co-occurring multimodal records, while the EM-Graph narrows the multi-year history to related episodes.

\noindent\textbf{Performance on List Recall Questions.}
\model (No-Forget) obtains 50.4\% on List Recall, compared with 39.3\% for HippoRAG2-SGM and 32.4\% for ATM-RAG-SGM. This category requires several relevant items, making it most aligned with ScrapMem's episodic organization. The 11.1-point margin indicates that the gain is concentrated where a single topically similar record is insufficient: a daily page retains temporally aligned multimodal evidence, and connected EM-Paths expose related events. Although the table evaluates their combined effect, the location of the gain supports the intended mechanism of preserving correlated evidence for multi-item answers.

\noindent\textbf{Where the Answering Gains Arise.}
The gain is not uniform across answer types. ScrapMem's Open-ended score (48.8\%) is only 0.4 points above Mem0-Agentic and Self-RAG-SGM, while its Number score (57.2\%) is below ATM-RAG-SGM (60.3\%). The overall improvement is therefore driven mainly by retrieval and List Recall, not a uniformly stronger answerer. Exact-value questions benefit less from event structure because success depends on reading one specific date, identifier, or quantity.

\noindent\textbf{Evidence Retrieval.}
\model (No-Forget) is the only evaluated system exceeding 70\% R@10, reaching 70.3\% versus 69.6\% for HippoRAG2-SGM and 68.7\% for Self-RAG-SGM and ATM-RAG-SGM. This gain is obtained without increasing the evidence budget, indicating better allocation of the same ten retrieval slots. Query nodes select related EM-Paths, whose source pages provide the fine-grained evidence used for answering (controlled analyses: Section~\ref{sec:controlled_main} and Appendix~\ref{sec:synthetic_emgraph_results}).

\noindent\textbf{Behavior after Optical Forgetting.}
The forgetting-enabled operating point retains 46.9\% Joint@10 and 66.1\% R@10, matching HippoRAG2-SGM on Joint@10. List Recall remains 45.3\%, above all baselines, whereas Number decreases from 57.2\% to 55.3\%. This asymmetry identifies what survives compression: EM-Paths can still locate an episode, but cannot reconstruct an illegible numerical detail. Optical Forgetting thus retains useful access structure while introducing a measurable evidence-fidelity cost (controlled analyses: Appendices~\ref{sec:synthetic_representation}--\ref{sec:synthetic_forgetting_results}).

\subsection{Sensitivity Analysis of Optical Forgetting}

\begin{table}[ht]
\centering
\small
\resizebox{\columnwidth}{!}{
\begin{tabular}{l ccc}
\toprule
\textbf{Configuration} & \textbf{JPEG Quality} & \textbf{Resolution Scale} & \textbf{Age Boundaries} \\
\midrule
No-Forget    & 100 / 100 / 100 & 1.0 / 1.0 / 1.0 & N/A \\
Very\_Soft             & 95 / 82 / 70    & 1.0 / 0.95 / 0.85 & 180 / 730d \\
Softer\_Old            & 90 / 75 / 60    & 1.0 / 0.90 / 0.80 & 180 / 730d \\
Timed-Gentle  & 90 / 70 / 40    & 1.0 / 0.85 / 0.60 & 180 / 730d \\
Boundary\_365          & 95 / 75 / 55    & 1.0 / 0.90 / 0.75 & 365 / 900d \\
\bottomrule
\end{tabular}
}
\caption{Hyperparameter settings for the forgetting strategies, ordered by compression strength. Each entry specifies the Recent/Mid-term/Old stages; age boundaries are measured in days.}
\label{tab:forget_configs}
\end{table}

\begin{figure}[t]
  \centering
  \includegraphics[width=1.0\linewidth]{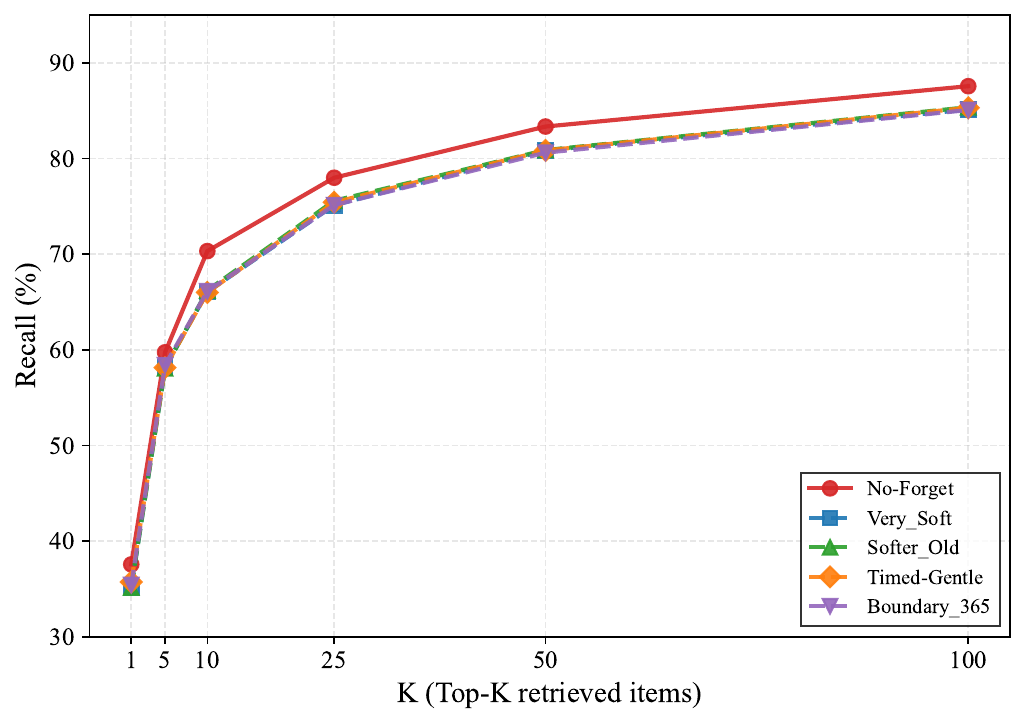}
  \Description{Recall at different retrieval depths for No-Forget and four Optical Forgetting configurations. The forgetting-enabled curves remain closely grouped, while No-Forget is consistently higher.}
  \caption{Retrieval performance (Recall@$K$) under varying Optical Forgetting intensities. The closely grouped forgetting curves show limited sensitivity to the evaluated hyperparameter configurations.}
  \label{fig:recall_sensitivity}
\end{figure}

Table~\ref{tab:forget_configs} changes JPEG quality, resolution scale, and age boundaries, while Figure~\ref{fig:recall_sensitivity} varies retrieval depth. All forgetting variants maintain R@10 around 66\% and follow closely grouped Recall@$K$ curves, whereas No-Forget remains consistently higher. Thus, the main retrieval loss appears when compression is introduced; within the evaluated range, the exact schedule has a secondary effect on locating the episode. Each policy changes several parameters together, so the result establishes robustness across complete operating policies without attributing the effect to one parameter.

\subsection{Effect of Canonical Threshold \texorpdfstring{$\theta_c$}{theta-c}}

\begin{figure}[t]
  \centering
  \includegraphics[width=1.0\linewidth]{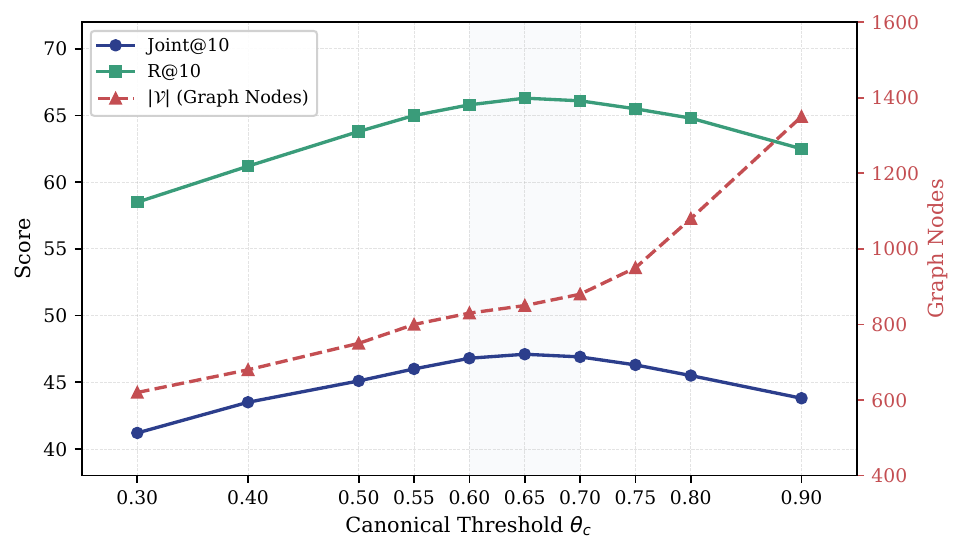}
  \Description{Joint at 10, Recall at 10, and EM-Graph size as the canonical node-merging threshold varies from 0.3 to 0.9. Retrieval metrics peak around 0.6 to 0.7, after which graph size increases sharply.}
  \caption{Sensitivity analysis of the canonical threshold $\theta_c$ in node merging (Algorithm~\ref{alg:node_merge} in Appendix~\ref{sec:appendix_algorithms}). Joint@10 and R@10 peak simultaneously within the shaded region $\theta_c \in [0.6, 0.7]$, while graph size $|\mathcal{V}|$ grows sharply beyond $\theta_c = 0.75$. The optimal interval achieves the best balance between redundancy suppression and retrieval recall.}
  \label{fig:tau_sensitivity}
\end{figure}

The canonical threshold $\theta_c$ trades graph compactness against semantic distinction. Under \textit{Timed-Gentle}, we sweep $\theta_c$ from 0.3 to 0.9. Figure~\ref{fig:tau_sensitivity} shows that Joint@10 and R@10 peak for $\theta_c\in[0.6,0.7]$ at approximately 850 nodes. Lower values merge distinctions needed for retrieval; above 0.75, graph size rises sharply without a corresponding gain, so the added nodes are largely redundant for the evaluated queries. We use $\theta_c=0.65$ based jointly on retrieval, answering, and graph size.

\subsection{Storage Efficiency}

\begin{figure}[t]
  \centering
  \includegraphics[width=1.0\linewidth]{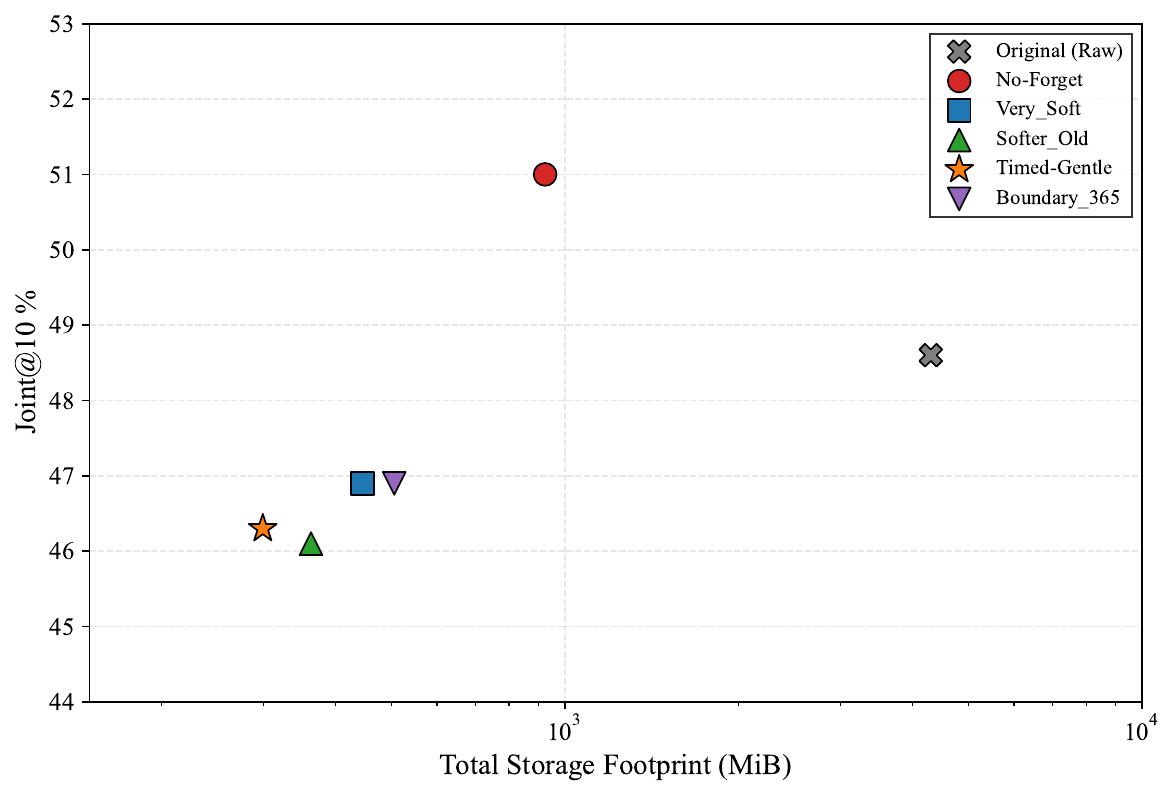}
  \Description{A log-scale scatter plot of persistent storage against Joint at 10. No-Forget has the highest score, while Timed-Gentle uses the least storage; the remaining forgetting configurations occupy nearby intermediate trade-off points.}
    \caption{\textbf{Storage--performance operating points on ATM-Bench.} The x-axis uses a logarithmic scale. \model (Timed-Gentle, orange star) reduces storage by 93.0\% relative to the raw-data baseline while retaining over 90\% of No-Forget performance (46.3\% vs.\ 51.0\% Joint@10).}
  \label{fig:storage_efficiency}
\end{figure}

Figure~\ref{fig:storage_efficiency} and Table~\ref{tab:storage_efficiency} separate daily consolidation from Optical Forgetting. No-Forget reduces the 4302.9 MiB corpus to 923.7 MiB while achieving 51.0\% Joint@10, showing that unified daily pages provide substantial first-stage savings without compression loss. \textit{Timed-Gentle} reduces the complete store to 299.5 MiB (93.0\%) with 46.3\% Joint@10. Relative to No-Forget, it removes another 624.2 MiB at a cost of 4.7 points.

\begin{table}[t]
\centering
\small
\resizebox{\columnwidth}{!}{
\begin{tabular}{l ccc c}
\toprule
\textbf{Configuration} & \textbf{Scrapbook} & \textbf{Total Storage} & \textbf{Reduction} & \textbf{Joint@10} \\
\midrule
Raw Corpus              & --- & 4302.9 MiB & --- & --- \\
\model (No-Forget)     & 896.0 MiB & 923.7 MiB & 78.5\% & \textbf{51.0\%} \\
\model (Timed-Gentle)  & 272.0 MiB & \textbf{299.5 MiB} & \textbf{93.0\%} & 46.3\% \\
\model (Softer\_Old)   & 338.0 MiB & 363.0 MiB & 91.6\% & 46.1\% \\
\model (Very\_Soft)    & 421.0 MiB & 446.0 MiB & 89.6\% & 46.9\% \\
\model (Boundary\_365) & 481.0 MiB & 506.0 MiB & 88.2\% & 46.9\% \\
\bottomrule
\end{tabular}
}
\caption{Storage efficiency. Total storage includes both Scrapbook Pages and the EM-Graph.}
\label{tab:storage_efficiency}
\end{table}

The reduction comes from the visual store, not a smaller retrieval index: Scrapbook Pages account for 896.0 of 923.7 MiB in No-Forget and 272.0 of 299.5 MiB in \textit{Timed-Gentle}. Optical Forgetting therefore targets the intended bottleneck while leaving graph storage nearly unchanged. Moreover, \textit{Very\_Soft} and \textit{Boundary\_365} both obtain 46.9\% Joint@10, although the latter uses 60.0 MiB more. The policies are deployment operating points, not a monotonic accuracy curve: \textit{Timed-Gentle} minimizes storage, while a softer policy recovers 0.6 Joint@10 points at additional cost.

\subsection{On-device Execution Analysis}
\label{sec:ondevice_analysis}

ScrapMem separates the latency-sensitive read path from asynchronous memory maintenance. During active interaction, the system performs query-node extraction, sparse Q-matrix matching, top-$k$ EM-Path retrieval, and answer generation. The Q-matrix avoids scanning every stored page: query nodes first identify candidate graph nodes, after which only their associated EM-Paths and source pages enter the answer context. Our laptop prototype uses an Intel Core i7-13700H CPU, an NVIDIA RTX 4070 GPU with 8\,GB VRAM, and 32\,GB RAM; both Qwen3-VL models use 4-bit quantization. On this prototype, the measured end-to-end latency of the interactive read path remains below 10 seconds.

Scrapbook Page perception, canonical node merging, EM-Graph construction and updating, and scheduled Optical Forgetting belong to the write and maintenance path. These operations update persistent memory without blocking the current response, so they can run during charging or device-idle intervals. Devices with limited perception capacity may optionally offload page perception or graph construction to a cloud--edge service while retaining the Scrapbook store, EM-Graph index, and retrieval policy locally. Runtime cost is dominated by model inference, whereas persistent memory is dominated by Scrapbook Page storage; sparse Q-matrix matching and EM-Graph maintenance remain outside the latency-critical path. The deployment policy determines whether raw pages or only derived representations may leave the device.

\subsection{Controlled Mechanism Study}
\label{sec:controlled_main}
\label{sec:synthetic_efficiency_cases}

We complement ATM-Bench with controlled six-month health and lifestyle histories for ten synthetic users, comprising more than 1,000 Scrapbook Pages and 400 queries across four tasks. All systems share Qwen2.5-3B and DeepSeek-OCR~\cite{wei2025deepseekocr}. GPT-4o scores personalization, temporal logic, causal accuracy, and selective forgetting from 1 to 5; Overall averages these four axes under a recency-aware filtering objective. Appendix~\ref{sec:controlled_synthetic} gives the protocol and full scores.

\begin{table}[t]
\centering
\caption{Overall scores on controlled histories under the recency-aware rubric (1--5). Best values are bold.}
\label{tab:synthetic_summary}
\small
\setlength{\tabcolsep}{6pt}
\renewcommand{\arraystretch}{1.12}
\begin{tabular}{>{\raggedright\arraybackslash}p{\dimexpr.5\linewidth-60pt\relax}*{4}{>{\raggedleft\arraybackslash}p{.125\linewidth}}}
\toprule
\textbf{Method} & \textbf{PP} & \textbf{CR} & \textbf{SE} & \textbf{PA} \\
\midrule
\rowcolor[RGB]{234,246,236}
No Memory & 1.79 & 2.05 & 1.93 & 2.00 \\
FIFO & 1.62 & 1.31 & 1.74 & 3.36 \\
NaiveRAG & \textbf{4.23} & 3.89 & \textbf{4.46} & 4.59 \\
Mem0 & 3.23 & 3.13 & 3.61 & 4.35 \\
GraphRAG & 3.38 & 3.13 & 3.21 & 4.00 \\
H$^2$Memory & 3.86 & 3.76 & 4.10 & 4.75 \\
\midrule
\rowcolor[RGB]{242,238,252}
w/o Forgetting & 3.06 & 3.61 & 3.81 & 4.45 \\
\rowcolor[RGB]{242,238,252}
w/o EM-Graph & 2.75 & 2.65 & 2.90 & 4.04 \\
\midrule
\rowcolor[RGB]{242,238,252}
\textbf{ScrapMem} & 4.05 & \textbf{3.92} & 4.24 & \textbf{4.79} \\
\bottomrule
\end{tabular}
\par\smallskip{\footnotesize\raggedright PP: Physiological Profiling; CR: Causal Reasoning; SE: Socio-psychological Evolution; PA: Preference Adaptation.\par}
\end{table}

Table~\ref{tab:synthetic_summary} separates the effects of Optical Forgetting and the EM-Graph in this controlled setting. ScrapMem exceeds \textit{w/o Forgetting} on all four tasks; removing the EM-Graph reduces Overall by 1.30, 1.27, 1.34, and 0.75 points. NaiveRAG leads in Physiological Profiling and Socio-psychological Evolution, while ScrapMem leads in Causal Reasoning and Preference Adaptation.

\begin{figure*}[t]
    \centering
    \includegraphics[width=0.88\linewidth]{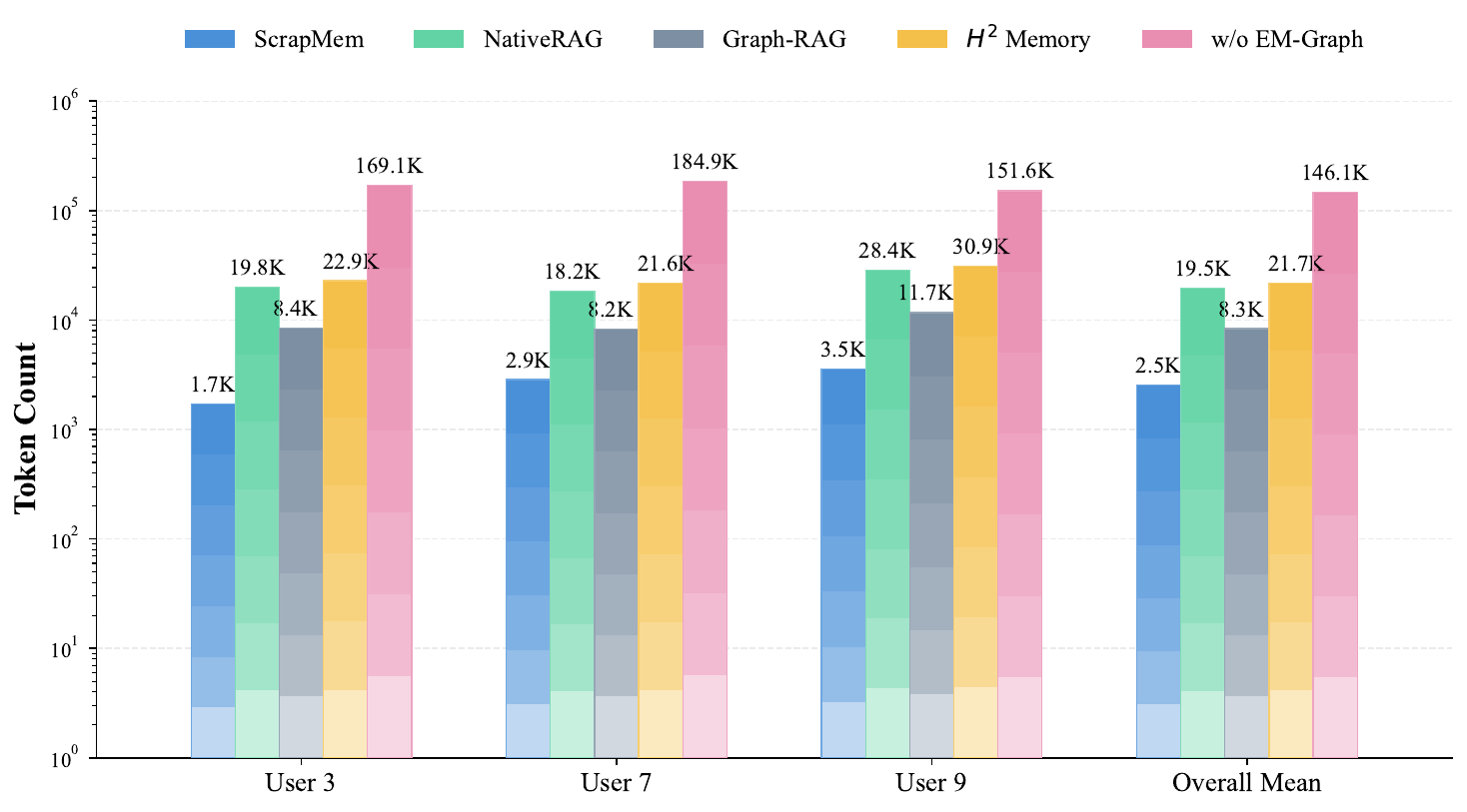}
    \caption{Cumulative context-token consumption over the controlled six-month histories. Tokens measure retrieved evidence context, not persistent storage. ``NativeRAG'' denotes NaiveRAG.}
    \Description{Cumulative retrieved-context tokens for ScrapMem and baselines over six months of controlled personal histories.}
    \label{fig:synthetic_token_comparison}
\end{figure*}

Figure~\ref{fig:synthetic_token_comparison} reports cumulative context consumption. ScrapMem uses a mean of 2,541.2 tokens versus 146,093 for \textit{w/o EM-Graph}, a reduction greater than 98\%. This measures retrieved evidence context, separately from persistent storage in Table~\ref{tab:storage_efficiency}.

\section{Discussion}

\subsection{Interpreting Optical Forgetting}

The small R@10 variation across forgetting schedules and the stronger List Recall result support the separation between EM-Graph access structure and visual evidence. The graph can still locate an episode after its page is compressed, but it is not a lossless substitute for that page. If a date, identifier, or small numerical value becomes unreadable, retrieval cannot reconstruct it. Optical Forgetting is therefore controlled lossy storage: it retains episodic access while accepting lower fidelity in aging evidence.

\subsection{Design Trade-offs}

Eviction makes a removed memory unavailable, while text summarization replaces source evidence with a generated abstraction whose omissions may be hard to inspect. Resolution scaling and JPEG compression retain a lower-fidelity version of the same page. The policy requires no learned salience predictor, has a predictable storage footprint, and keeps the remaining evidence inspectable.

The fixed three-stage age schedule also provides predictable storage and maintenance timing, and the sensitivity results show that performance is not tied to one narrow setting. However, recency is only a proxy for future utility: an old medical record may remain important, while a recent notification may not. Importance-aware forgetting could improve this trade-off but introduces another prediction problem, so adaptive policies remain future work.

\subsection{Temporal Unit of Consolidation}

One calendar day preserves temporal proximity without a separate segmentation model and gives every page an unambiguous age. It is an organizational boundary, not a one-event assumption: semantic nodes and EM-Paths separate events within a page and connect dependencies across days. Event-adaptive or capacity-based pages may better handle unusually dense histories but are outside the current scope.

\subsection{Limitations}

The evaluation establishes ScrapMem's storage--performance advantage across a public multimodal benchmark and a controlled mechanism study. The ATM-Bench evaluation measures the complete framework; further component-level ablations can quantify the individual contributions of page consolidation and EM-Graph retrieval. ATM-Bench covers emails, images, and videos; audio, continuous sensors, and application-specific records may require different page and forgetting policies. The threshold $\theta_c$ may also need recalibration under different entity densities or embedding distributions. Further deployment studies will characterize energy consumption, peak memory usage, and per-component latency across edge platforms.

\section{Conclusion}
ScrapMem addresses long-term personalized agent memory by separating multimodal evidence storage from structured retrieval. It consolidates each day's records into a Scrapbook Page, organizes their semantics into causal-temporal EM-Paths, and compresses aging pages through Optical Forgetting. The EM-Graph thereby retains stable access paths even when older evidence is stored at lower fidelity.

On ATM-Bench, ScrapMem (No-Forget) establishes a new non-oracle state of the art with 51.0\% Joint@10 and 70.3\% Recall@10; its largest answering gain occurs on List Recall questions requiring multiple records. Optical Forgetting reduces persistent storage by up to 93.0\% while retaining over 90\% of No-Forget performance, establishing an inspectable storage--fidelity trade-off. The controlled evaluation further identifies the resolution boundary and component effects (Appendix~\ref{sec:synthetic_takeaway}). Together, daily multimodal consolidation, structured episodic retrieval, and asynchronous maintenance provide a practical foundation for storage-constrained on-device agents. Future work includes importance-aware forgetting and device-specific energy and runtime optimization.

\begin{acks}
Supported by the ``111 Center'' (No.~B26023).
\end{acks}


\bibliographystyle{ACM-Reference-Format}
\clearpage
\bibliography{sample-base}

\appendix
\section{Implementation Details}
\label{sec:appendix}

\subsection{Experimental Settings}
\label{sec:appendix_setup}
The ATM-Bench experiments follow the official baseline configurations, with identical answer prompts and decoding settings across systems. A-Mem uses Piled/Linked organization, Mem0 uses Agentic/Plain answering, HippoRAG2 uses Linked retrieval, and ATM-RAG uses Piled retrieval. Each video contributes at most eight frames. Under \textit{Timed-Gentle}, the canonical threshold sweep is $\theta_c\in\{0.3,0.4,0.5,0.55,0.6,0.65,0.7,0.75,0.8,0.9\}$; the reported configuration uses $0.65$.

\subsection{Prompt Specifications}
\label{sec:appendix_prompts}
Record-side extraction returns a JSON \texttt{semantic\_nodes} array of 3--8 concise lowercase phrases covering entities, objects, actions, events, and time mentions, excluding field names. Query-side extraction applies the same categories to the question, returning 3--10 phrases without stop words. Both use temperature 0; output limits are 256 and 200 tokens, respectively, with a 30-second query timeout.

Page perception reads visible text, timestamps, identifiers, and non-text visual evidence, returning \texttt{ocr\_text}, \texttt{visual\_summary}, and \texttt{salient\_items}. Path consolidation receives the date and perception text and returns at most ten \texttt{semantic\_nodes} plus one concise \texttt{em\_path} sentence. These specifications summarize the extraction and consolidation instructions.

\subsection{Graph Update Algorithms}
\label{sec:appendix_algorithms}
Algorithms~\ref{alg:node_merge} and~\ref{alg:pruning} give canonical merging and forgetting-driven pruning. The encoder $\phi$ is all-MiniLM-L6-v2; $Q_{ij}$ records whether path $P_i$ contains node $v_j$.

\begin{algorithm}[b]
\small
\caption{Semantic Node Merging and Canonical Update}
\label{alg:node_merge}
\Require{Page nodes $\mathcal{V}_t$, global nodes $\mathcal{V}$, encoder $\phi$, threshold $\theta_c$}
\ForEach{$v_i\in\mathcal{V}_t$}{
    $v^*\leftarrow\varnothing$\;
    \If{$\mathcal{V}\neq\varnothing$}{
        $v^*\leftarrow\arg\max_{v_j\in\mathcal{V}}\cos(\phi(v_i),\phi(v_j))$\;
    }
    \eIf{$v^*\neq\varnothing$ and $\cos(\phi(v_i),\phi(v^*))\geq\theta_c$}{
        Merge $v_i$ into $v^*$ and update its center/metadata\;
    }{
        $\mathcal{V}\leftarrow\mathcal{V}\cup\{v_i\}$\;
    }
}
\Return $\mathcal{V}$\;
\end{algorithm}

\begin{algorithm}[b]
\small
\caption{EM-Graph Pruning via Optical Forgetting}
\label{alg:pruning}
\Require{Graph $\mathcal{G}=(\mathcal{V},\mathcal{P})$, matrix $Q$, original nodes $\mathcal{V}_t$, re-perceived nodes $\mathcal{V}_t^{(\tau)}$}
\ForEach{$v_j\in\mathcal{V}_t\setminus\mathcal{V}_t^{(\tau)}$}{
    \ForEach{$P_i$ with $Q_{ij}=1$}{
        $Q_{ij}\leftarrow0$\;
        \If{$P_i$ fails semantic coherence}{
            $\mathcal{P}\leftarrow\mathcal{P}\setminus\{P_i\}$\;
        }
    }
    \If{$\sum_i Q_{ij}=0$}{
        $\mathcal{V}\leftarrow\mathcal{V}\setminus\{v_j\}$\;
        $\mathcal{G}\leftarrow\mathrm{Purge}(\mathcal{G},v_j)$\;
    }
}
\Return $(\mathcal{V},\mathcal{P}),Q$\;
\end{algorithm}

\section{Controlled Study Details}
\label{sec:controlled_synthetic}

\subsection{Data and Evaluation Protocol}
\label{sec:synthetic_data}
\label{sec:synthetic_protocol}
Ten virtual users aged 25--45 have demographic, occupational, health, family, personality, and preference profiles. Big Five traits use Low/Medium/High levels. Qwen-Plus generates approximately 100 events per user from June to December 2025. Qwen-Image-Max (Wanx) renders photographs; programmatic templates render application screenshots and medical documents. The resulting histories contain more than 1,000 Scrapbook Pages.

Each user has ten queries in each of four tasks: Physiological Profiling, Causal Reasoning, Socio-psychological Evolution, and Preference Adaptation (400 queries total). All systems use Qwen2.5-3B for answering and DeepSeek-OCR for perception. The compared baselines include FIFO, NaiveRAG~\cite{lewis2020rag}, Mem0~\cite{chhikara2025mem0}, GraphRAG~\cite{edge2024graphrag}, and H$^2$Memory~\cite{huang2025mempal}.

GPT-4o assigns 1--5 scores for Personalization Influence (PI), Temporal Logic (TL), Causal Accuracy (CA), and Forgetting Intelligence (FI); Overall is their unweighted mean, averaged across queries. The rubric rewards user-specific evidence, temporal distinction, causal consistency, and selective use of relevant early context; it penalizes indiscriminate memory dumping and hallucinated causal relations. Recent high-fidelity evidence is prioritized, with early memories treated as background. This recency-aware objective complements the evidence-grounded ATM-Bench evaluation.

\subsection{Representation Fidelity}
\label{sec:synthetic_representation}
Table~\ref{tab:synthetic_representation} measures per-page Medical Entity Recall (ME-Recall), the fraction of annotated medical entities recovered, and Noise Reduction Ratio (NRR), the fraction of recognized outputs judged valid. At 512/256/128 px, ME-Recall is 0.9289/0.8600/0.1695: fine detail is lost sharply at 128 px. These stages apply to the controlled study; ATM-Bench uses the independent schedules in Table~\ref{tab:forget_configs}. Graph retrieval cannot recover details that are no longer legible.

\par\medskip\noindent\begin{minipage}{\linewidth}
\centering
\captionof{table}{Per-page representation fidelity. Text uses text tokens; image rows use vision tokens. Disk space is in MB.}
\label{tab:synthetic_representation}
\small
\setlength{\tabcolsep}{0pt}
\renewcommand{\arraystretch}{1.0}
\begin{tabular*}{\linewidth}{@{\extracolsep{\fill}}lrrrr@{}}
\toprule
\textbf{Resolution} & \textbf{Disk} & \textbf{Tokens} & \textbf{ME-Recall} & \textbf{NRR} \\
\midrule
Text & 0.5853 & 168 & 1.0000 & 1.0000 \\
\addlinespace[2pt]
1024 px & 153.3690 & 256 & 0.9969 & 0.4854 \\
640 px & 1.6389 & 100 & 0.9997 & 0.4804 \\
512 px & 0.9479 & 64 & 0.9289 & 0.5466 \\
256 px & 0.3197 & 16 & 0.8600 & 0.2981 \\
128 px & 0.1101 & 4 & 0.1695 & 0.0634 \\
64 px & 0.0401 & 1 & 0.1262 & 0.0944 \\
\bottomrule
\end{tabular*}
\end{minipage}\par\medskip

\subsection{Full Results}
\label{sec:synthetic_forgetting_results}
\label{sec:synthetic_emgraph_results}
\label{sec:synthetic_takeaway}
Tables~\ref{tab:synthetic_forgetting} and~\ref{tab:synthetic_emgraph} retain the full metric breakdown behind Table~\ref{tab:synthetic_summary}. ScrapMem improves FI over \textit{w/o Forgetting} by 1.31, 0.51, 0.44, and 0.47 points. NaiveRAG has higher CA in every task, and H$^2$Memory also exceeds ScrapMem on CA throughout, illustrating the importance of detailed evidence for causal answers. The controlled results characterize selective historical retrieval; Figure~\ref{fig:synthetic_token_comparison} separately reports its context cost.

\par\medskip\noindent\begin{minipage}{\linewidth}
\centering
\captionof{table}{Controlled forgetting comparison (1--5). Best values per task are bold.}
\label{tab:synthetic_forgetting}
\footnotesize
\setlength{\tabcolsep}{0pt}
\renewcommand{\arraystretch}{1.04}
\begin{tabular*}{\linewidth}{@{\extracolsep{\fill}}lrrrrr@{}}
\toprule
\textbf{Method} & \textbf{PI} & \textbf{TL} & \textbf{CA} & \textbf{FI} & \textbf{Overall} \\
\midrule
\multicolumn{6}{@{}l}{\textbf{a) Physiological Profiling}} \\
\addlinespace[2pt]
FIFO & 1.62 & 1.62 & 1.62 & 1.62 & 1.62 \\
NaiveRAG & 3.78 & \textbf{4.45} & \textbf{4.56} & 4.13 & \textbf{4.23} \\
Mem0 & 2.85 & 3.28 & 3.52 & 3.25 & 3.23 \\
w/o Forgetting & 2.78 & 3.23 & 3.25 & 2.97 & 3.06 \\
\textbf{ScrapMem} & \textbf{3.90} & 4.15 & 3.85 & \textbf{4.28} & 4.05 \\
\midrule
\multicolumn{6}{@{}l}{\textbf{b) Causal Reasoning}} \\
\addlinespace[2pt]
FIFO & 1.12 & 1.12 & 1.16 & 1.84 & 1.31 \\
NaiveRAG & 3.55 & \textbf{3.95} & \textbf{4.15} & 3.89 & 3.89 \\
Mem0 & 2.69 & 3.15 & 3.01 & 3.66 & 3.13 \\
w/o Forgetting & 3.20 & 3.89 & 3.53 & 3.81 & 3.61 \\
\textbf{ScrapMem} & \textbf{3.72} & 3.85 & 3.78 & \textbf{4.32} & \textbf{3.92} \\
\midrule
\multicolumn{6}{@{}l}{\textbf{c) Socio-psychological Evolution}} \\
\addlinespace[2pt]
FIFO & 1.74 & 1.74 & 1.74 & 1.74 & 1.74 \\
NaiveRAG & \textbf{4.06} & \textbf{4.62} & \textbf{4.72} & \textbf{4.44} & \textbf{4.46} \\
Mem0 & 3.44 & 3.69 & 3.82 & 3.48 & 3.61 \\
w/o Forgetting & 3.61 & 3.84 & 3.86 & 3.91 & 3.81 \\
\textbf{ScrapMem} & 3.95 & 4.42 & 4.25 & 4.35 & 4.24 \\
\midrule
\multicolumn{6}{@{}l}{\textbf{d) Preference Adaptation}} \\
\addlinespace[2pt]
FIFO & 3.12 & 3.10 & 3.67 & 3.56 & 3.36 \\
NaiveRAG & 4.44 & 4.67 & \textbf{4.71} & 4.54 & 4.59 \\
Mem0 & 4.16 & 4.42 & 4.42 & 4.38 & 4.35 \\
w/o Forgetting & 4.38 & 4.38 & 4.58 & 4.45 & 4.45 \\
\textbf{ScrapMem} & \textbf{4.75} & \textbf{4.85} & 4.62 & \textbf{4.92} & \textbf{4.79} \\
\bottomrule
\end{tabular*}
\end{minipage}\par\medskip

\par\medskip\noindent\begin{minipage}{\linewidth}
\centering
\captionof{table}{Controlled EM-Graph ablation and memory baselines (1--5). Best values per task are bold.}
\label{tab:synthetic_emgraph}
\footnotesize
\setlength{\tabcolsep}{0pt}
\renewcommand{\arraystretch}{1.04}
\begin{tabular*}{\linewidth}{@{\extracolsep{\fill}}lrrrrr@{}}
\toprule
\textbf{Method} & \textbf{PI} & \textbf{TL} & \textbf{CA} & \textbf{FI} & \textbf{Overall} \\
\midrule
\multicolumn{6}{@{}l}{\textbf{a) Physiological Profiling}} \\
\addlinespace[2pt]
No Memory & 1.79 & 1.79 & 1.79 & 1.79 & 1.79 \\
GraphRAG & 2.99 & 3.49 & 3.79 & 3.23 & 3.38 \\
w/o EM-Graph & 2.48 & 2.80 & 2.87 & 2.84 & 2.75 \\
H$^2$Memory & 3.51 & 4.06 & \textbf{4.37} & 3.51 & 3.86 \\
\textbf{ScrapMem} & \textbf{3.90} & \textbf{4.15} & 3.85 & \textbf{4.28} & \textbf{4.05} \\
\midrule
\multicolumn{6}{@{}l}{\textbf{b) Causal Reasoning}} \\
\addlinespace[2pt]
No Memory & 2.03 & 2.05 & 2.07 & 2.03 & 2.05 \\
GraphRAG & 2.76 & 3.07 & 3.37 & 3.30 & 3.13 \\
w/o EM-Graph & 2.40 & 2.45 & 2.48 & 3.25 & 2.65 \\
H$^2$Memory & 3.35 & \textbf{4.17} & \textbf{4.15} & 3.38 & 3.76 \\
\textbf{ScrapMem} & \textbf{3.72} & 3.85 & 3.78 & \textbf{4.32} & \textbf{3.92} \\
\midrule
\multicolumn{6}{@{}l}{\textbf{c) Socio-psychological Evolution}} \\
\addlinespace[2pt]
No Memory & 1.91 & 1.94 & 1.95 & 1.91 & 1.93 \\
GraphRAG & 3.00 & 3.24 & 3.48 & 3.13 & 3.21 \\
w/o EM-Graph & 2.80 & 3.04 & 2.89 & 2.87 & 2.90 \\
H$^2$Memory & 3.73 & 4.36 & \textbf{4.57} & 3.73 & 4.10 \\
\textbf{ScrapMem} & \textbf{3.95} & \textbf{4.42} & 4.25 & \textbf{4.35} & \textbf{4.24} \\
\midrule
\multicolumn{6}{@{}l}{\textbf{d) Preference Adaptation}} \\
\addlinespace[2pt]
No Memory & 1.85 & 2.07 & 2.20 & 1.87 & 2.00 \\
GraphRAG & 3.82 & 4.09 & 4.11 & 3.96 & 4.00 \\
w/o EM-Graph & 4.02 & 4.05 & 4.05 & 4.04 & 4.04 \\
H$^2$Memory & 4.69 & 4.79 & \textbf{4.83} & 4.69 & 4.75 \\
\textbf{ScrapMem} & \textbf{4.75} & \textbf{4.85} & 4.62 & \textbf{4.92} & \textbf{4.79} \\
\bottomrule
\end{tabular*}
\end{minipage}\par\medskip

\end{document}